\documentclass[a4paper,fleqn]{cas-dc}

\usepackage[numbers]{natbib}
\usepackage{times}
\usepackage{indentfirst}
\usepackage{microtype}      % microtypography
\usepackage{soul}           % text highlighting
\usepackage{xcolor}         % colors

\usepackage{amsmath}        % math symbols
\usepackage{amsthm}
\usepackage{amsfonts}       % blackboard math symbols
\usepackage{bbding}         % checkmarks and other symbols
\usepackage{pifont}         % additional symbols

\usepackage{graphicx}       % graphics inclusion
\usepackage{booktabs}       % professional-quality tables
\usepackage{colortbl}       % table colors
\usepackage{multirow}       % multirow cells in tables
\usepackage{wrapfig}        % wrapping text around figures
\usepackage{float}          % improved float placement
\usepackage{subcaption}     % subfigures
\usepackage{overpic}        % overlay pictures with text
\usepackage{rotating}       % rotating objects, like tables and figures

\usepackage{algorithm}
\usepackage{algorithmic}

\usepackage{url}
\usepackage{listings}       % code listings

\usepackage{tikz,pgfplots}  % drawing and plotting

\usepackage{caption}

\usepackage{nicefrac}       %

\usepackage[capitalize]{cleveref}
\crefname{section}{Sec.}{Secs.}
\Crefname{section}{Section}{Sections}
\Crefname{table}{Table}{Tables}
\crefname{table}{Tab.}{Tabs.}

\def\tsc#1{\csdef{#1}{\textsc{\lowercase{#1}}\xspace}}
\tsc{WGM}
\tsc{QE}
\pgfplotsset{compat=1.18}
\begin{document}
\let\WriteBookmarks\relax
\def\floatpagepagefraction{1}
\def\textpagefraction{.001}

% Short title
\shorttitle{Hybrid Mask Generation for Infrared Small Target Detection with Single-Point Supervision}

% Short author
\shortauthors{Weijie He, et.al.}

% \title{Hybrid Mask Generation for Infrared Small Target Detection with Single-Point Supervision}  
\title [mode = title]{Hybrid Mask Generation for Infrared Small Target Detection with Single-Point Supervision}
% \tnotemark[1,2]

% \tnotemark[1] 
% \tnotetext[1]{} 

\author[1]{Weijie He}
\credit{Conceptualization, Investigation, Methodology, Experiment, Software, Validation, Visualization, Writing - Original Draft, Writing - Review \& Editing, Data Curation}
\author[1]{Mushui Liu}
\credit{Writing - Original Draft, Writing - Review \& Editing, Data Curation}
\author[1]{Yunlong Yu}
\credit{Conceptualization, Writing - Original Draft, Validation, Resources}
\cormark[1]
\ead{yuyunlong@zju.edu.cn}
% Address/affiliation
\affiliation[1]{
    organization={College of Information Science \& Electronic Engineering},
    addressline={Zhejiang University}, 
    city={Hangzhou},
    % citysep={}, % Uncomment if no comma needed between city and postcode
    postcode={310027}, 
    % state={},
    country={China}}
            
% \author[2]{}%[]

% Address/affiliation
% \affiliation[2]{organization={},
%             addressline={}, 
%             city={},
% %          citysep={}, % Uncomment if no comma needed between city and postcode
%             postcode={}, 
%             state={},
%             country={}}

% Corresponding author text
\cortext[1]{Corresponding author}

% Footnote text
% \fntext[1]{}

% For a title note without a number/mark
%\nonumnote{}

% Here goes the abstract
\begin{abstract}
Single-frame infrared small target (SIRST) detection poses a significant challenge due to the requirement to discern minute targets amidst complex infrared background clutter. In this paper, we focus on a weakly-supervised paradigm to obtain high-quality pseudo masks from the point-level annotation by integrating a novel learning-free method with the hybrid of the learning-based method. The learning-free method adheres to a sequential process, progressing from a point annotation to the bounding box that encompasses the target, and subsequently to detailed pseudo masks, while the hybrid is achieved through filtering out false alarms and retrieving missed detections in the network's prediction to provide a reliable supplement for learning-free masks. The experimental results show that our learning-free method generates pseudo masks with an average Intersection over Union (IoU) that is 4.3\% higher than the second-best learning-free competitor across three datasets, while the hybrid learning-based method further enhances the quality of pseudo masks, achieving an additional average IoU increase of 3.4\%.
\end{abstract}

% Use if graphical abstract is present
%\begin{graphicalabstract}
%\includegraphics{}
%\end{graphicalabstract}

% Research highlights
% \begin{highlights}
% \item 
% \item 
% \item 
% \end{highlights}

% Keywords
% Each keyword is seperated by \sep
\let\printorcid\relax
\maketitle

\section{Introduction}
Infrared Small Target Detection (IRSTD), a critical task aimed at isolating minute objects from complex infrared backgrounds, holds immense potential for diverse applications such as traffic management and public safety \cite{Rawat2020ReviewOR, empowering2021, wu2022ship}. In real-world scenarios, these targets often occupy a minute fraction of pixels and exhibit low signal intensities, rendering them highly susceptible to being obscured within the background imagery, posing significant challenges for accurate detection.

Existing IRSTD frameworks are mostly based on the deep neural network in a fully supervised way \cite{MDvsFA, Zhang2022ExploringFC, yang2024EFLNet, Zhang2024IRPruneDetEI}, which requires a large amount of labeled data. However, the annotation process for pixel-level masks is labor-intensive and challenging to process accurately due to the limited color and texture information, as well as the fuzzy boundaries of small targets in infrared (IR) imagery \cite{cheng2022pointly,li2021fully}. To alleviate the annotation burden, weakly-supervised methods \cite{bilen2016weakly, bearman2016whats, wei2017stc}, which utilize inexact annotations, have emerged as an effective and prominent solution. 

\begin{figure*}[t]
    \centering
    \begin{overpic}[width=1.0\linewidth]{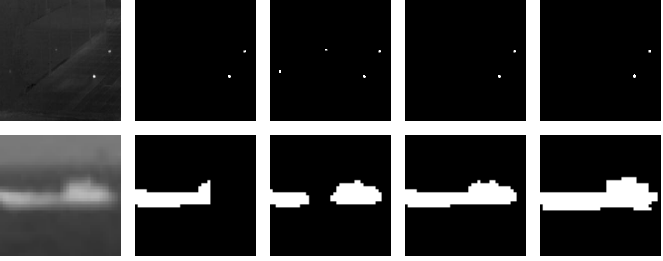}
    \put(-2,26){\rotatebox{90}{IRSTD-1K}}
     \put(-2,5){\rotatebox{90}{NUDT-SIRST}}
    \put(2,-2){(a) Infrared Image}
    \put(20,-2){(b) Learning-free Mask}
    \put(40,-2){(c) Learning-based Mask}
    \put(64,-2){(d) Hybrid Mask}
    \put(83,-2){(e) Ground-Truth}
    \end{overpic}
    \vspace{0.01cm}
    \caption{Illustration samples from the {IRSTD-1K} \cite{ISNet} and NUDT-SIRST \cite{DNANet} datasets showcase different mask generation techniques. The learning-free methods are often constrained by cropping size, leading to missed detections in pixel-level outside the cropping area. The deep-learning model is not constrained by spatial limitations, while inevitably facing false alarms or missed detections in target-level. The hybrid method, combining the strengths of both learning-free and deep-learning models, obtains high-quality masks.}
    \label{fig:barfigure}
\end{figure*}

The key to the weakly-supervised IRSTD paradigm is to obtain high-quality masks from the point-level annotations. Currently, existing approaches are roughly categorized into two groups: learning-based methods and learning-free methods. The learning-based method, e.g., LESPS \cite{ying2023mapping}, progressively expands point labels into mask labels by utilizing a label evolution mechanism within the deep-learning framework. However, the model's predictions may be highly uncertain due to the unreliable supervision and the continuous iterations may result in the problem of error accumulation. The learning-free methods \cite{li2023monte,li2024levelset, kou2024mcgc} develop masks through handcrafted algorithms that differentiate the background from the target based on pixel values. To prevent extensive false detections on the overall image when encountering small targets that are too similar to the background, the learning-free methods employ cropping sizes \cite{li2023monte, li2024levelset} or distance attenuation matrix \cite{kou2024mcgc} to confine the assessment of small target pixels to a designated area, which inevitably results in missed detections for target pixels located outside of this specified region, as illustrated in Fig.~\ref{fig:barfigure}. In contrast, the learning-based methods do not impose such spatial constraints, enabling them to process pixel annotations beyond the predefined region. To tackle this issue, MCLC \cite{li2023monte} further takes the size of each target as the prior information, which crops a region of a specific size centered around the point label, within which the pseudo mask is generated. 

To investigate the impact of cropping size on learning-free algorithms in the absence of size prior information, we take the cropping size as a hyperparameter for all targets within the same dataset. The experimental results revealed that, under the condition of lacking size prior information, the quality of the pseudo masks exhibits significant variability depending on the choice of cropping size. When the cropping size deviates considerably from the actual size of the targets, there is a significant decline in the quality of the pseudo masks.

To alleviate the reliance on prior information about target size, we propose a hybrid approach that takes the merits of both learning-based and learning-free methods to obtain high-quality pseudo masks from the point-level annotation. Specifically, we first present a learning-free method called Point-to-Mask Generation (PMG) that adheres to a sequential process, progressing from point labels to bounding boxes, and subsequently to detailed initial pseudo masks. PMG can adaptively estimate the size of each target based on the observation that the pixel differences within the target region and background region are smaller than the pixel differences between the target and background regions. Even if the preset box is significantly larger than the target, PMG could effectively filter out the vast majority of the background, thereby obtaining an approximate shape and size of the target. The initial pseudo-masks are then employed to supervise the training of the neural network, directing its predictions to converge toward the pseudo-masks. The predictions of the neural network, as the learning-based masks are then integrated with masks obtained from the learning-free method through False Alarm Filtering and Missed Detection Retrieving, to obtain hybrid masks. While the learning-free method helps correct the network's predictions, the neural network assists in completing pixels outside the cropping-size confine encountered by learning-free methods in turn. This complementary nature enables us to obtain hybrid masks with enhanced quality, as shown in Fig.~\ref{fig:barfigure}(d).

To summarize, our key highlights are:
\begin{itemize}
    \item We propose a size-aware, learning-free method named Point-to-Mask Generation (PMG), which decomposes the mask generation process into a sequential pipeline involving points, bounding boxes, and masks, enabling adaptive estimation of the size of each target.
    \item We present a novel hybrid mask generation approach to leverage the complementary strengths of both deep-learning and learning-free techniques to provide comprehensive and reliable mask supervision for IRSTD.
    \item Experimental results on three SIRST datasets demonstrate that the models trained with our hybrid masks set a new benchmark under single-point supervision.   
\end{itemize}

\section{Related work}

% In this paper, we also leverage the benefits of multi-modal alignment and the generalization ability of CLIP. By fine-tuning the CLIP model in limited data regimes, we investigate how the model can adapt its knowledge and generalize to perform well in this particular challenging scenario.

\subsection{Infrared Small Target Detection}
Early traditional approaches to infrared small target detection (IRSTD) primarily relied on handcrafted features and statistical methodologies. These methods often focused on enhancing the visibility of targets against backgrounds through filtering-based \cite{d1999maxmean, Rivest1996DetectionOD}, local contrast-based \cite{chen2014local, wang2012infrared} or low-rank-based \cite{gao2013lowrank, Gao2013InfraredPM} techniques. However, in complex real-world scenarios, these methods lacked sufficient generalization capability. With the development of deep learning, the research focus has shifted from the traditional paradigm to the data-driven paradigm based on deep learning. For example, ACM\cite{ACM} introduced an asymmetric contextual modulation module to effectively encode high-level contextual information while preserving finer details of targets. Meanwhile, DNANet\cite{DNANet} presented a dense nested attention network to comprehensively exploit contextual cues of small targets. Furthermore, SCTransNet\cite{yuan2024sctransnet} proposed a spatial-channel cross-transformer network to reinforce the semantic differences between targets and clutter at multiple levels. Rather than focusing on designing advanced feature extraction or fusion modules, MSHNet\cite{liu2024cvpr} introduced a novel scale and location-sensitive loss function for IRSTD, which helps detectors better distinguish objects with varying scales and locations. TCI-Former\cite{chen2024tci-former} was the first to introduce heat conduction theories into the IRSTD network design, establishing a connection between the spatial and temporal information of pixel values during the IRSTD process. However, these methods are trained in a fully supervised manner based on pixel-level annotated images, which incurs significant human annotation costs to obtain large-scale pixel-level labeled data for these data-driven models. EDGSP\cite{yuan2024beyond} proposed to enhance the IRSTD network with the single-point prompt during inference. Notably, unlike single-point supervised IRSTD, EDGSP still falls under the category of fully supervised methods, as it utilizes ground truth masks during the training process and additionally provides single-point labels during inference.

\subsection{Weakly Supervised Segmentation with Points} In the field of weakly-supervised segmentation, Bearman\cite{bearman2016whats} first employed single-point supervision in the network training process. Following this, Laradji\cite{laradji2020proposal} constructed a network with two branches to predict the location of the target and group pixels with similar embeddings, to obtain the identified target mask. Building on the concept of point supervision, PDML\cite{qian2019sceneparsing} achieved semantic scene parsing under multi-point supervision by optimizing the intra- and inter-category embedding feature consistency among the annotated points. Additionally, recent works have further extended single-point supervision to the domain of IRSTD. For example, LESPS\cite{ying2023mapping} discovered the phenomenon of mapping degradation during deep neural network training, and proposed a label evolution framework that gradually expands the initial point-level labels to provide better supervision quality. Similarly, MCLC\cite{li2023monte} utilized a Monte Carlo linear clustering method to predict high-quality pseudo masks from point labels for training supervision while also relying on the size prior information of targets for better mask generation. Additionally, COM\cite{li2024levelset} designed a dedicated energy functional based on the intensity expectation difference between the areas around the target, realizing the prediction of pseudo masks from point labels. Existing methods struggle to obtain high-quality masks for guiding model training when prior information is unavailable. In this paper, we propose a solution that leverages the strengths of both handcrafted algorithms and neural networks to generate robust and high-quality pseudo-masks.

\begin{figure*}[ht]
    \centering
    \includegraphics[width=\textwidth]{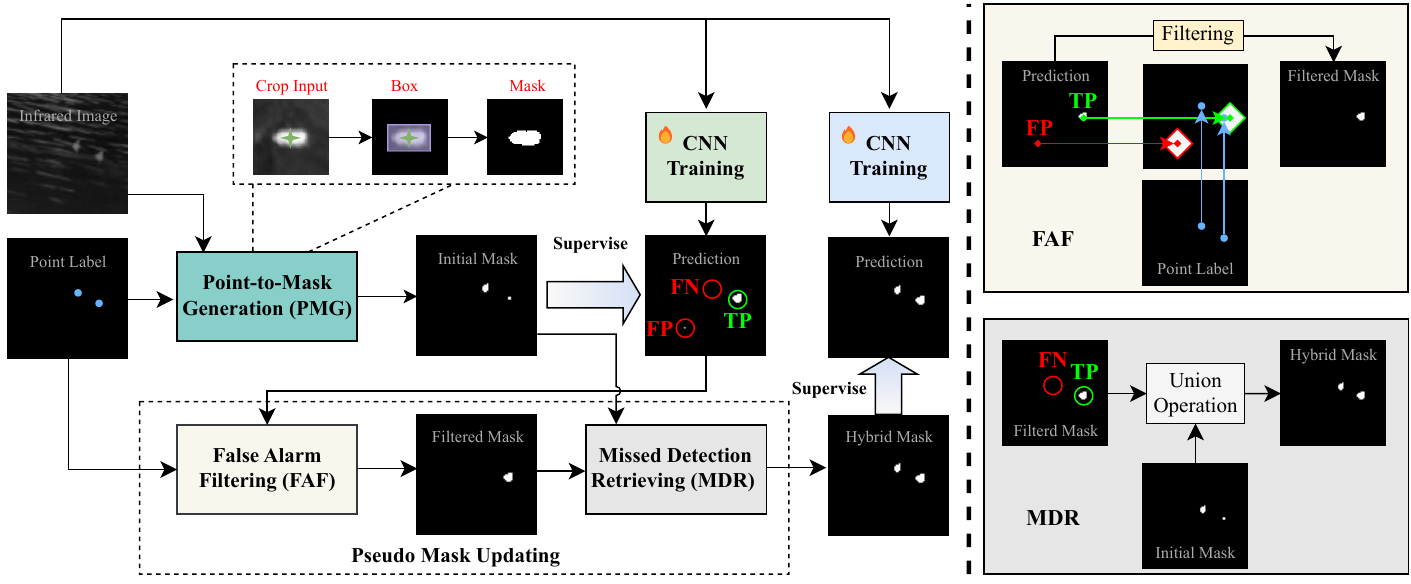}
    \caption{Illustration of the proposed pseudo-mask generation process. The PMG module is utilized to generate initial masks for the supervision of the IRSTD model. The Pseudo Mask Updating module, composed of FAF and MDR components, is used to combine the initial mask with the model's predictions for higher-quality pseudo masks. }
    \label{fig:pipeline}
\end{figure*}

\section{Method}

As illustrated in Fig.~\ref{fig:pipeline}, our method initially generates high-quality pseudo masks from point-level annotations, which are then utilized for subsequent deep model training. The pseudo-masks generation consists of two stages. First, the initial masks are obtained with a learning-free method, i.e., Point-to-Mask Generation (PMG), to provide initial supervision for the training of the IRSTD model. PMG processes the point annotation, transforming it into a bounding box that encompasses the target, and ultimately into the initial mask. In the second stage, a Pseudo-Mask-Updating strategy composed of a False Alarm Filtering (FAF) module and a Missed Detection Retrieving (MDR) module is proposed to perform the correction of the model's predictions and the complementary fusion with the initial masks, obtaining the final hybrid masks. These hybrid masks are then used as the supervision to retrain the IRSTD model.

\subsection{Point-to-Mask Generation}
The point-to-mask generation strategy entails a point-to-box step and a box-to-mask step to obtain initial pseudo-masks from point-level annotations. 

\subsubsection{Point-to-Box (P2B)}
\label{sec:p2b}
The point-to-box step involves predicting a bounding box that encapsulates the target. Based on the observation that the pixel differences within the target and background regions are significantly smaller than inter-region differences, we categorize the pixels surrounding the point label into three distinct regions: the target, the transition, and the background. Then, we establish a threshold using the pixels located at the intersection of the background and transition regions to locate the boundary between the target and pure surroundings in each direction. Finally, we derive the bounding box by integrating the boundary predictions from these four directions.

\begin{figure}[t]
    \centering
    \begin{overpic}[width=0.9\linewidth]{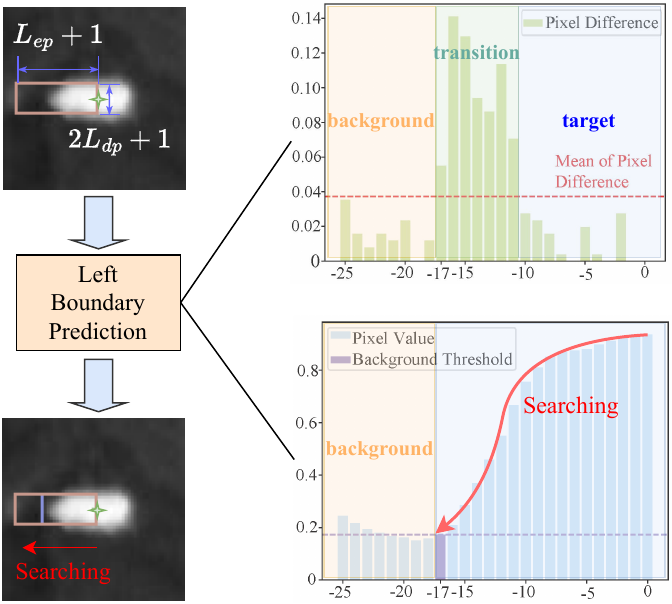} 
    \put(60,45){\footnotesize (a) Pixel Differences}
    \put(55,-3){\footnotesize (b) Vertical average vector}
    \end{overpic}
    \caption{The illustration of obtaining the left boundary of the bounding box from the point label.}
    \label{fig:bounding}
\end{figure}

Fig.~\ref{fig:bounding} shows the determination of the left boundary. A rectangular area of interest in the infrared image is defined by using the point label as the reference point $u_0$ and expanding it $L_{ep}$ pixels to the left, $L_{dp}$ pixels in upward and downward directions. Notably, $2L_{ep}+1$ serves as the cropping size for each target in our method. Within this rectangular area, we compute the vertical maximum vector and vertical average vector by identifying the maximum and average pixel values along the vertical dimension, respectively. Then, we calculate the absolute differences between adjacent elements in the vertical maximum vector, yielding the pixel differences $D$ along the prediction direction. As depicted in Fig.~\ref{fig:bounding} (a), while the overall pixel differences are small, there is a noticeable distribution with the middle region exhibiting larger values compared to the two ends. Based on this observation, we calculate the mean of $D$ to establish a difference threshold, which allows us to categorize the area into three regions: the target, transition, and background regions. Once we identify the position of the boundary where the transition region meets the background region, we designate the corresponding value in the vertical average vector as the background threshold, as shown in Fig.~\ref{fig:bounding}(b). Starting from the point annotation in the vertical average vector, we search leftward until we find a position where the value crosses the background threshold. This position is then identified as the left boundary. 

Similarly, we predict the remaining right, upward, and downward boundaries, thereby determining the bounding box that encompasses the target. 

\subsubsection{Box-to-Mask (B2M)}
\label{sec:b2m}
After obtaining the bounding box, we proceed to evaluate the label of each pixel within it. Specifically, we establish a pixel threshold $\sigma$ to distinguish the background and the target, where the threshold is calculated by incorporating both the mean pixel value within the bounding box and the pixel value at the position of $u_0$. Incorporating the pixel value at $u_0$ enhances the mask quality and helps mitigate the effects of any potential errors in the bounding box prediction. Specifically, the threshold $\sigma$ is obtained with:
\begin{equation}
     \sigma = \alpha * I(u_0) + (1-\alpha) * \frac{1}{N} \sum_{u \in \textit{U}} I(u)
\end{equation}
where $I(u)$ denotes the pixel value at the spatial location $u$ in the infrared image. The set $U$ encompasses all pixels within the bounding box and $N$ is the pixel number in $U$.

Pixels with values exceeding $\sigma$ are more likely to belong to the target, while those with lower values are more likely to be part of the background. To quantify this, we normalize the pixel values to obtain target probabilities within $U$. Specifically, we distribute pixel values lower than $\sigma$ into a target probability $P(u)$ range of $0\mbox{-}0.5$ and pixel values higher than $\sigma$ into a target probability $P(u)$ range of $0.5\mbox{-}1$. This can be expressed as follows:
\begin{equation}
     P(x,y) = \left\{
     \begin{array}{lc}
        \dfrac{I(u)-Min}{\sigma-Min}*0.5 & I(u)<=\sigma \\
        1-\dfrac{Max-I(u)}{Max-\sigma}*0.5 & I(u)>\sigma,
     \end{array}
     \right.
\end{equation}
where $Min$ and $Max$ are the minimum and maximum pixel value within $U$. 

Considering the distinct characteristics of target and background pixels in different directions, we extend our analysis beyond the global perspective to create a global probability map $P_g$ within $U$ and perform direction-specific probability predictions. Specifically, we divide $U$ into separate subregions along four cardinal directions (upward, downward, leftward, and rightward) from the reference point $u_0$. Each subregion is then normalized to produce a corresponding probability map.

To this end, the probability maps for the upward and downward directions are concatenated to achieve vertical bidirectional normalization, resulting in $P_v$. Similarly, the probability maps for the leftward and rightward directions are concatenated to achieve horizontal bidirectional normalization, yielding $P_h$. The final target probability map $P_f$ is derived by averaging the probabilities across three distinct probability maps. Following this aggregation, we apply a threshold of 0.5 to $P_f$, converting it into a binary pseudo mask $PM$. That is:
\begin{equation}
    P_f(u) = \frac{P_g(u)+P_v(u)+P_h(u)}{3},
\end{equation}
\begin{equation}
    PM(u) = \left\{
    \begin{array}{ll}
        1 & P_f(u)>=0.5 \\
        0 & P_f(u)<0.5
    \end{array}
    \right.
\end{equation}

After obtaining the initial pseudo masks for the training samples, these samples with masks can subsequently be employed to train neural networks.

\subsection{Pseudo Mask Updating}
Due to the limitations of cropping size or distance attenuation matrix, handcrafted algorithms often tend to generate pseudo masks within a small cropping region, neglecting pixels outside it. Therefore, it is necessary to utilize the predictions of the neural network to complete the pixels outside the region. However, it is important to note that the predictions of the neural network are also not entirely reliable, and may suffer from false alarms or missed detections for small targets. Therefore, to achieve complementarity between the learning-free algorithm and neural network capabilities, we propose utilizing the Pseudo Mask Updating module to correct the predictions of the neural network and assist in refining the initial masks to obtain hybrid masks. As depicted in Fig.~\ref{fig:pipeline}, the updating process encompasses two key modules: a False Alarm Filtering (FAF) module and a Missed Detection Retrieving (MDR) module.

\subsubsection{False Alarm Filtering (FAF)}
\label{sec:FAF}
To mitigate the false alarm issue in the predictions of the learning-based model, we design an effective filtering strategy. We first calculate the centroids of each connected component within the predictions and then compare these centroids with the point labels. A connected component in prediction is classified as a false alarm if the $L_1$ distance between its centroid and every point from the point labels exceeds $r$, where $r$ is set to 30. Subsequently, we set the pixels of the connected components classified as false alarms to zero in order for filtering. After filtering out all false alarms in the predictions, we obtain the filtered mask.

\subsubsection{Missed Detection Retrieving (MDR)}
To address potential missed detections in the filtered masks and to simultaneously leverage the strengths of both the learning-free algorithm and the deep neural network, we compute the union of the initial mask and the filtered mask to obtain the final hybrid mask. In this way, the targets that are neglected with the learning-based method and the pixels outside the cropping region with the learning-free method would be reconsidered.

\definecolor{gray1}{gray}{.92}
\begin{table*}
\centering
\scriptsize
    \resizebox{0.99\linewidth}{!}{
    \setlength{\tabcolsep}{2.4mm}
    \begin{tabular}{l|l|ccc|ccc|ccc|ccc}
        \toprule
        \multirow{2}*{Models}&\multirow{2}*{Methods}&\multicolumn{3}{c|}{IRSTD-1K}&\multicolumn{3}{c|}{NUAA-SIRST} &\multicolumn{3}{c|}{NUDT-SIRST} &\multicolumn{3}{c}{Mean}\\\cline{3-14}
&&$\rm{IoU\uparrow}$&$\rm{P_d\uparrow}$&$\rm{F_a\downarrow}$&$\rm{IoU\uparrow}$&$\rm{P_d\uparrow}$&$\rm{F_a\downarrow}$&$\rm{IoU\uparrow}$&$\rm{P_d\uparrow}$&$\rm{F_a\downarrow}$&$\rm{IoU\uparrow}$&$\rm{P_d\uparrow}$&$\rm{F_a\downarrow}$\\
        \midrule
        \multirow{5}*{ACM} &\cellcolor{gray1}Full Supervision  \cite{ACM} &\cellcolor{gray1}60.97&\cellcolor{gray1}90.58&\cellcolor{gray1}21.78&\cellcolor{gray1}70.33&\cellcolor{gray1}93.91&\cellcolor{gray1}3.73&\cellcolor{gray1}67.08&\cellcolor{gray1}95.97&\cellcolor{gray1}10.18&\cellcolor{gray1}66.13&\cellcolor{gray1}93.49&\cellcolor{gray1}11.90 \\
        &LESPS \cite{ying2023mapping}&41.44&88.89&60.46&49.23&89.35&40.95&42.09&91.11&38.24&44.25&89.78&46.55 \\
        &COM \cite{li2024levelset}&53.18&91.92&88.16&57.76&85.93&27.44&36.63&64.87&42.54&49.19&80.91&52.71 \\
        &MCLC \cite{li2023monte}&55.60&93.60&\textbf{49.31}&67.31&91.63&\textbf{21.54}&56.44&87.51&43.00&59.78&90.91&\textbf{37.95} \\
        &HMG (Ours)&\textbf{59.66}&\textbf{94.28}&63.35&\textbf{70.33}&\textbf{93.54}&30.94&\textbf{60.55}&\textbf{92.06}&\textbf{36.97}&\textbf{63.51}&\textbf{93.29}&43.75 \\
        \midrule
        \multirow{6}*{DNANet}&\cellcolor{gray1}Full Supervision \cite{DNANet} &\cellcolor{gray1}68.44&\cellcolor{gray1}94.77&\cellcolor{gray1}8.81&\cellcolor{gray1}76.24&\cellcolor{gray1}97.71&\cellcolor{gray1}12.80&\cellcolor{gray1}86.36&\cellcolor{gray1}97.39&\cellcolor{gray1}6.90&\cellcolor{gray1}77.01&\cellcolor{gray1}96.62&\cellcolor{gray1}9.50 \\
        &LESPS \cite{ying2023mapping}&52.09&88.88&\textbf{16.09}&61.95&92.02&18.17&57.99&94.71&26.45&57.34&91.87&\textbf{20.24} \\
        &COM \cite{li2024levelset}&49.21&86.87&35.43&58.73&88.59&35.47&40.41&71.11&88.77&49.45&82.19&53.22 \\
        &LELCM \cite{yang2024label}&55.23&87.15&22.67&58.71&92.26&38.49&58.30&89.46&\textbf{22.43}&57.41&89.62&27.86 \\
        &MCLC \cite{li2023monte}&57.37&93.27&26.00&71.74&94.30&31.01&68.60&94.23&45.41&65.90&93.60&34.14 \\
        &HMG (Ours)&\textbf{61.21}&\textbf{94.95}&20.65&\textbf{72.48}&\textbf{95.06}&\textbf{10.22}&\textbf{73.21}&\textbf{95.56}&29.92&\textbf{68.97}&\textbf{95.19}&20.26 \\
        \bottomrule
    \end{tabular}}
        \caption{$\rm{IoU}$ ($\times 10^{-2}$), $\rm {P_d}$ ($\times 10^{-2}$) and $\rm{F_a}$($\times 10^{-6}$) values of different methods achieved on {IRSTD-1K} \cite{ISNet}, {NUAA-SIRST \cite{ALCNet}}, and {NUDT-SIRST} \cite{DNANet} datasets. "Full Supervision" represent models trained under pixel-level supervision and other methods are under single-point supervision. For a fair comparison, the results of \cite{li2023monte} were obtained by reproducing experiments without size prior information, using the best cropping size selected for each dataset. The best performance among the single-point supervised methods is indicated in \textbf{bold}.}
    \label{tab:mainresult}
\end{table*}

\section{Experiments}

\subsection{Experiment Settings}
\subsubsection{Dataset Details}
We conduct evaluations on three popular datasets: NUAA-SIRST \cite{ALCNet}, IRSTD-1K \cite{ISNet}, and NUDT-SIRST \cite{DNANet} and follow the same splits and data augmentation as LESPS\cite{ying2023mapping} to partition three datasets into their respective training sets, test sets and process images during training. By default, the point labels in the training set are derived from the centroids of small infrared targets. Additionally, following the settings of LESPS, we also explored the scenario of using coarse centroid in Section 1.

\subsubsection{Implement Details}
$L_{ep}$ was set to 10 for NUDT-SIRST and 25 for the remaining datasets. $L_{dp}$ and $\alpha$ were set to 4 and 0.15 for all datasets, respectively. The networks were trained with Soft-IoU loss function and optimized with Adam \cite{Kingma2014AdamAM}, with a batch size of 8. The initial learning rate was set to 1e-3 and was reduced with the CosineAnnealingLR scheduler. The training epochs of the two stages were set to 250 and 1250, respectively, to maintain the same epochs in total as MCLC \cite{li2023monte}. Without specification, the default SIRST model used in experiments is DNANet \cite{DNANet}. All models were implemented in PyTorch \cite{Paszke2019PyTorchAI} on a PC with an Nvidia GeForce 3090 GPU. In the experiment, two pixel-level metrics (i.e., Intersection over Union (IoU) and False alarm rate ($\rm F_a$)) and one instance-level metric (Probability of detection ($\rm P_d$)) are employed for performance evaluation.

\subsection{Performance Comparison}
\subsubsection{Networks' Performance on Test Sets}
Two classic SIRST methods, namely ACM \cite{ACM} and DNA-Net \cite{DNANet}, serve as the training models in our study. From the results shown in Table~\ref{tab:mainresult}, we observe that our method performs very competitively on all three datasets. Compared with the state-of-the-art method \cite{li2023monte}, our method achieves an approximately 3\% improvement in the IoU metric and over 1.6\% improvement in the $\rm P_d$ metric on the average across all three datasets and both models. Notably, when using ACM as the training model, our method performs close to the fully supervised counterpart in terms of both IoU and $\rm P_d$ metrics.

Although our method doesn't show an advantage in the $\rm F_a$ metric, we believe this does not detract from the overall superiority of our approach. In the area of IRSTD, Fa denotes the pixel-level false detection rate, and the values are shown in the order of $10^{-6}$. In the results of DNANet\cite{DNANet}, our method achieved a value nearly identical to that of the best competitor. And in the results of ACM\cite{ACM}, our method's $\rm F_a$ ranked second, with a difference of approximately $6\times10^{-6}$ from the first place, which corresponds to a difference of merely about 1.57 pixels in a 512×512 image. Moreover, $\rm F_a$ is inherently sensitive to the threshold used during mask binarization while IoU avoids threshold-related biases. Recent competition LimitIRSTD \cite{li2024first} adopt a weighted sum of IoU and Pd as the primary metric, requiring only that Fa remain below a reasonable threshold. Our HMG has $\rm F_a$ values comparable in magnitude to other methods, while with significant improvements in IoU and Pd, which are critical for real-world applications.

\begin{table}[ht]
\centering
    \resizebox{\linewidth}{!}{
    \setlength{\tabcolsep}{2.4mm}
    \begin{tabular}{l|c|c|c|c}
        \toprule
         Method & IRSTD-1k	&NUAA-SIRST	& NUDT-SIRST &Mean \\
         \midrule
         COM \cite{li2024levelset} &32.88 &18.46	&7.43 &19.59 \\
         SAM \cite{SAM} &59.32 & 69.26 &44.40 &57.66 \\
         MCGC \cite{kou2024mcgc} & 59.31 &69.55 &68.33 &65.73 \\
         MCLC \cite{li2023monte} & 62.23 &70.59 &65.26 &66.03 \\
         PMG (Ours) &65.79 &74.22 &70.98	&70.33 \\
        \bottomrule
    \end{tabular}}
    \caption{IoU (\%) of generated pseudo mask on three datasets in the training stage. For a fair comparison, the results of \cite{li2023monte} were obtained using the best cropping size as the prior information for each dataset.}
    \label{tab:initialization}
\end{table}

\subsubsection{Quality of Initial Pseudo Masks on Training Sets}
In this experiment, we compare our PMG method with other learning-free pseudo-mask generation techniques and generalized segmentation model SAM's \cite{SAM} biggest version: sam\_vit\_h. As shown in Table~\ref{tab:initialization}, our PMG method exhibits superior performance, achieving an average IoU improvement of over 4\% compared to the second-best methods across the three datasets. Notably, compared to other models, SAM's performance does not demonstrate superiority, which may be attributed to the gap in modality and target size. The high-quality pseudo-masks produced by our method would contribute to the subsequent training of deep models.

\begin{table}[ht]
\centering
    \resizebox{\linewidth}{!}{
    \setlength{\tabcolsep}{2.4mm}
    \begin{tabular}{l|c|c|c|c}
        \toprule
         Method & IRSTD-1k	&NUAA-SIRST	& NUDT-SIRST &Mean \\
         \midrule
         Prior+MCLC \cite{li2023monte} &  68.80	&76.30	&60.20	&68.43 \\
         MCLC \cite{li2023monte} &  62.23	&70.59	&65.26	&66.03 \\
         P2B+MCLC &  65.09	&73.37	&66.99	&68.48 \\
         P2B+B2M (PMG) & 	65.79	&74.22	&70.98	&70.33\\
        \bottomrule
    \end{tabular}}
    \caption{IoU (\%) of generated pseudo mask of MCLC under different conditions and our PMG.}
    \label{tab:crop}
\end{table}

\subsection{Ablation Study}
\subsubsection{Effects of Point-to-Mask Generation} 
\label{sec:ep2b}
Our proposed Point-to-Mask Generation consists of two steps: Point-to-Box(P2B) and Box-to-Mask(B2M). We will subsequently analyze the challenges encountered by the state-of-the-art competitor MCLC\cite{li2023monte} in generating pseudo masks and progressively introduce our Point-to-Box and Box-to-Mask steps to demonstrate the effectiveness of our approach.

\textbf{Effects of Point-to-Box.} In the training set, MCLC\cite{li2023monte} not only records the single-point coordinates of small targets but also additionally captures their true sizes as prior information. This prior information is used to condition the cropping of candidate regions for small targets in infrared images, thereby generating high-quality pseudo masks, which brings more annotation cost than pure single-point supervision. However, in the absence of size prior information, the cropping size of candidate regions is treated as a hyperparameter in the dataset, and the quality of pseudo masks generated by MCLC is significantly compromised, as shown in Table~\ref{tab:crop}. To analyze the impact of cropping size, we measured the average IoU of pseudo masks in three datasets for targets of different sizes under various cropping conditions. The classification criteria for target sizes align with the size prior information utilized in MCLC. \textbf{Point} refers to targets with pixels not exceeding 9, \textbf{Spot} refers to targets with pixels greater than 9 but not exceeding 81, and \textbf{Extended} refers to targets with pixels greater than 81.

\begin{figure}[t]
    \centering
    \begin{overpic}[width=\linewidth]{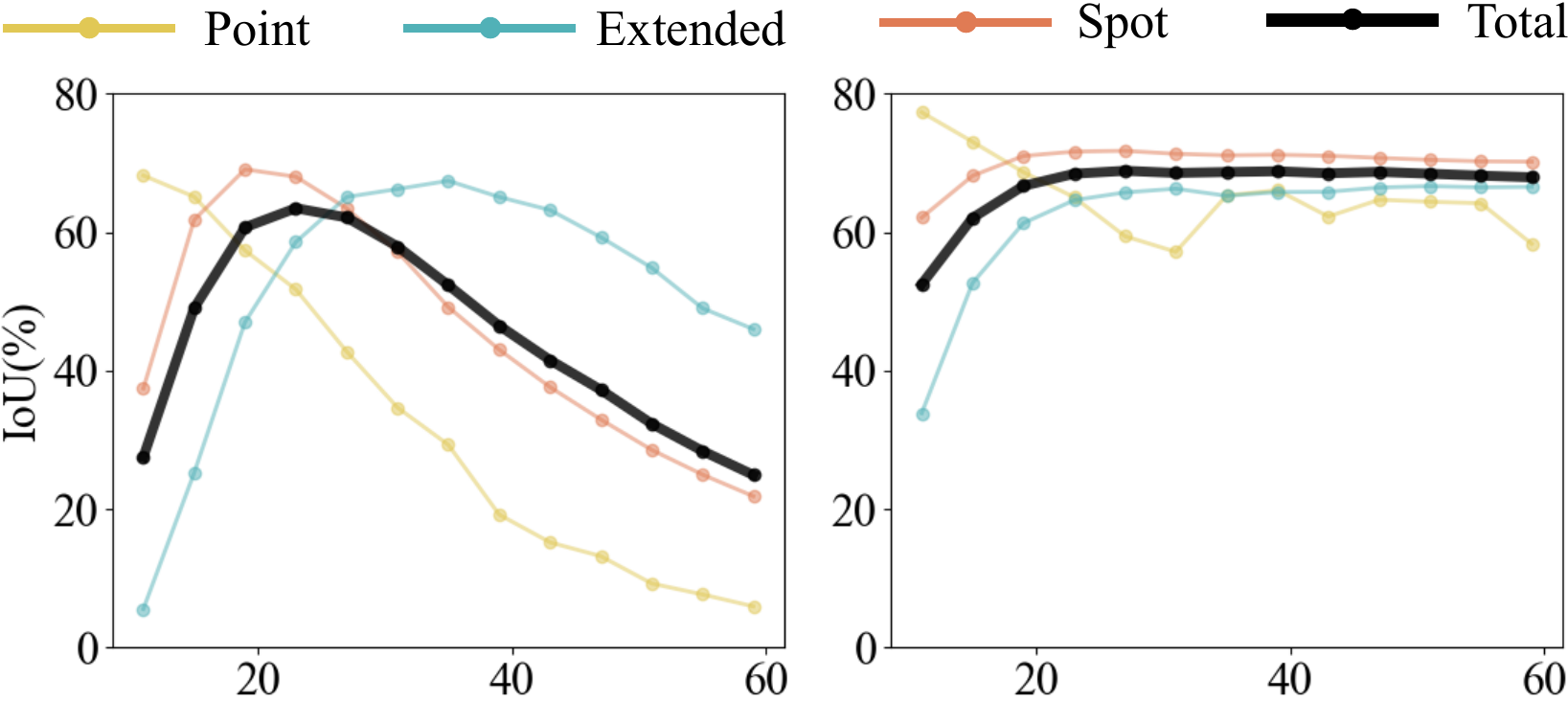} 
    \put(20,-3){\footnotesize (a) MCLC \cite{li2023monte}}
    \put(70,-3){\footnotesize (b) PMG (Ours)}
    \end{overpic}
    \vspace{0.01cm}
    \caption{Impacts of cropping size on average of three datasets for the quality of pseudo masks with different sizes of small targets for MCLC \cite{li2023monte} and our PMG.}
    \label{fig: crop_target_type}
\end{figure}

From Fig.~\ref{fig: crop_target_type}(a), it is evident that the performance of MCLC is sensitive to the cropping size. \textbf{Spot} targets achieve peak performance at a cropping size of 19, and \textbf{Extended} targets reach their maximum at a cropping size of 35, while the quality of the \textbf{Point} target has consistently declined since cropping size of 11. Additionally, significant deviations from the optimal cropping size for each target type lead to a drastic deterioration in the quality of the pseudo-masks produced by MCLC undergoes a drastic deterioration. Consequently, without prior knowledge of the target size, the consistency of MCLC's pseudo-mask quality cannot be guaranteed when dealing with infrared targets of unknown sizes.
Our Point-to-Box can eliminate most of the background within a large candidate region to obtain a more precise target bounding box, even in the absence of size prior information. Therefore, we conducted experiments to integrate Point-to-Box with MCLC, enabling MCLC to generate pseudo-masks within the refined target bounding box. From Table~\ref{tab:crop} we observe stable performance enhancements on all three datasets and the average IoU of the method is comparable to that of MCLC utilizing prior information, which demonstrates that Point-to-Box step can be flexibly integrated into other pseudo-mask generation methods, effectively minimizing the reliance on size-related prior information and enhancing the quality of pseudo-masks for small targets of unknown sizes.

\textbf{Effects of Box-to-Mask.} By replacing MCLC with Box-to-Mask to form our current Point-to-Mask Generation, we observed an additional 1.85\% improvement in average IoU across three datasets, thereby validating the effectiveness of Box-to-Mask step. From Fig.~\ref{fig: crop_target_type}(b), our PMG consistently generates high-quality pseudo-mask for a variety of small targets, even when the cropping size significantly exceeds the actual target size. This robustness is attributed to our Point-to-Box strategy, which allows for adaptive size estimation of small targets. Additionally, the peak quality of pseudo masks generated by PMG for various target types consistently surpasses that of MCLC, further underscoring the superiority of our proposed method. It is noteworthy that the IoU of the \textbf{Point} line initially decreases with increasing cropping size but experiences a dramatic rise around the size of 33, and we will provide an explanation for this phenomenon in the ~\ref{sec:drop}.

\begin{table*}
\centering

\scriptsize
    \resizebox{0.99\linewidth}{!}{
    \setlength{\tabcolsep}{2.4mm}
    \begin{tabular}{l|ccc|ccc|ccc|ccc}
        \toprule
        \multirow{2}*{Mask Type}&\multicolumn{3}{c|}{IRSTD-1K}&\multicolumn{3}{c|}{NUAA-SIRST} &\multicolumn{3}{c|}{NUDT-SIRST} &\multicolumn{3}{c}{Mean}\\\cline{2-13}&$\rm{IoU\uparrow}$&$\rm{P_d\uparrow}$&$\rm{F_a\downarrow}$&$\rm{IoU\uparrow}$&$\rm{P_d\uparrow}$&$\rm{F_a\downarrow}$&$\rm{IoU\uparrow}$&$\rm{P_d\uparrow}$&$\rm{F_a\downarrow}$&$\rm{IoU\uparrow}$&$\rm{P_d\uparrow}$&$\rm{F_a\downarrow}$\\
        \midrule
        Initial Mask&65.79&\textbf{98.58}&\textbf{7.66}&74.22&98.15&13.16&70.98&98.37&15.53&70.33&98.37&12.12 \\
        Prediction Mask&62.07&93.47&27.12&75.64&97.41&18.04&73.35&96.41&17.40&70.35&95.76&20.85 \\
        Filtered Mask&65.60&93.47&9.25&75.73&97.41&17.27&74.52&96.41&\textbf{4.99}&71.95&95.76&10.50 \\
        Hybrid Mask&\textbf{69.43}&98.49&8.65&\textbf{76.96}&\textbf{98.52}&\textbf{4.32}&\textbf{74.86}&\textbf{99.24}&8.49&\textbf{73.75}&\textbf{98.75}&\textbf{7.15} \\
        \bottomrule
    \end{tabular}}
    \caption{Impacts (\%) of pseudo mask updating on three datasets. The best result is
in \textbf{bold}.}
    \label{tab:updating}
\end{table*}

\subsubsection{Effects of Pseudo Mask Updating}
Table~\ref{tab:updating} presents the quality of different masks during the Pseudo Mask Updating process. By comparing the initial mask with the prediction mask, it can be observed that after training in Stage 1, the IRSTD model can generate prediction masks with an IoU comparable to that of the initial mask produced with PMG. However, the Probability of detection ($\rm P_d$) is significantly lower than that of the initial mask. This discrepancy occurs because the initial mask is generated using point labels that indicate the locations of small targets, whereas the IRSTD model lacks access to point label information during prediction, resulting in some missed detections.

\textbf{Effects of FAF.} By comparing the prediction mask with the filtered mask, it is evident that the application of the False Alarm Filtering module, which utilizes point labels to filter out false alarms from the prediction, leads to a significant reduction in the False alarm rate ($\rm F_a$), thereby contributing to an improvement in terms of IoU.

\textbf{Effects of MDR.} After processing through the Missed Detection Retrieving module (Hybrid Mask), the resulting hybrid mask outperforms all previously generated masks across all three metrics, achieving the $\rm P_d$ close to 100\% once again. This indicates that the hybrid mask generation module effectively combines the advantages of both learning-free and learning-based methods to produce high-quality masks that provide valuable supervisory information for the model. Notably, under conditions where point labels are available, both the initial mask and the hybrid mask achieve a $\rm P_d$ close to 100\%, and we will provide an explanation for this phenomenon in the ~\ref{sec:100}.

\begin{table*}
\centering
\scriptsize
    \resizebox{0.99\linewidth}{!}{
    \setlength{\tabcolsep}{2.4mm}
    \begin{tabular}{l|ccc|ccc|ccc|ccc}
        \toprule
        \multirow{2}*{Paradigm}&\multicolumn{3}{c|}{IRSTD-1K}&\multicolumn{3}{c|}{NUAA-SIRST} &\multicolumn{3}{c|}{NUDT-SIRST} &\multicolumn{3}{c}{Mean}\\\cline{2-13}&$\rm{IoU\uparrow}$&$\rm{P_d\uparrow}$&$\rm{F_a\downarrow}$&$\rm{IoU\uparrow}$&$\rm{P_d\uparrow}$&$\rm{F_a\downarrow}$&$\rm{IoU\uparrow}$&$\rm{P_d\uparrow}$&$\rm{F_a\downarrow}$&$\rm{IoU\uparrow}$&$\rm{P_d\uparrow}$&$\rm{F_a\downarrow}$\\
        \midrule
        HMG&61.21&94.95&20.65&72.48&95.06&10.22&73.21&95.56&29.92&68.97&95.19&20.26 \\
        HMG+Coarse Centroid&61.94&93.60&17.23&72.16&95.44&12.01&73.03&95.87&21.79&69.04&94.97&17.01 \\
        HMG+Point Prompt&72.69&98.65&3.32&75.60&98.10&13.93&75.37&99.05&17.33&74.55&98.60&11.53 \\
        \bottomrule
    \end{tabular}}
    \caption{Extended experiments on real-world scenarios of our method.}
    \label{tab:extended}
\end{table*}

\begin{figure*}[ht]
    \centering
    \begin{overpic}[width=\linewidth]{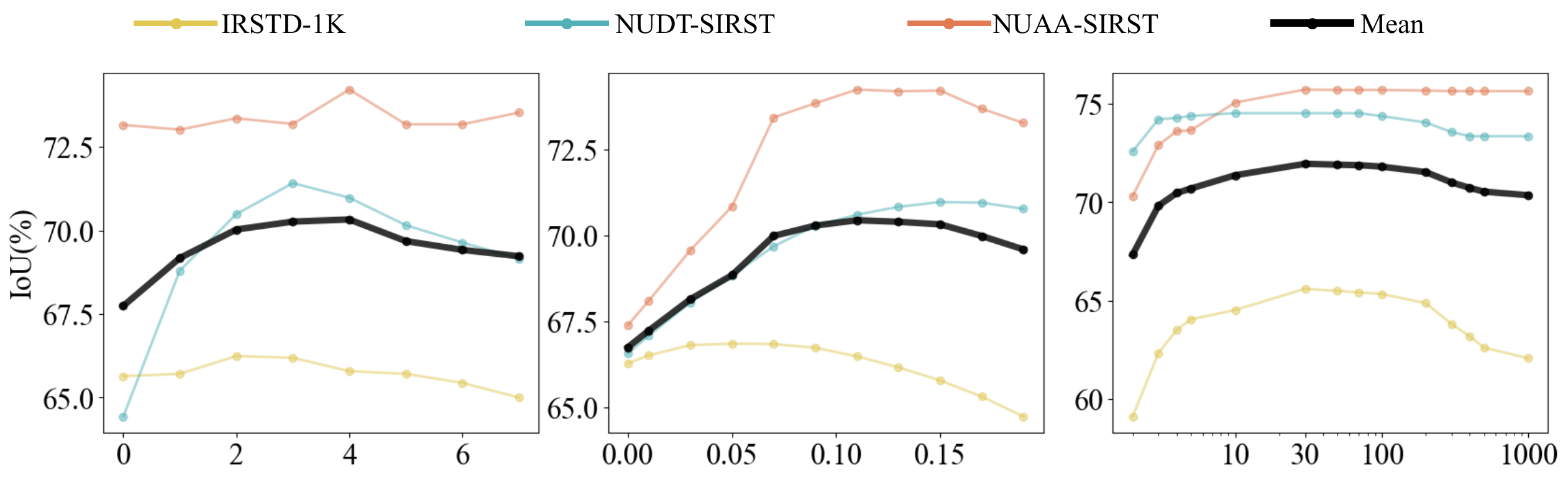}
    \put(17,-1){(a) $L_{dp}$}
    \put(50,-1){(b) $\alpha$}
    \put(85,-1){(c) $r$}
    \end{overpic}
    % \vspace{0.01cm}
    \caption{Impacts of $L_{dp}$ and $\alpha$ for the quality of initial pseudo masks on three datasets.}
    \label{fig: ablation}
\end{figure*}

\subsubsection{Impact of Hyperparameters}
Since $2L_{ep}+1$ can be regarded as the cropping size for each target in our method and the impacts of cropping size have already been discussed in \ref{sec:ep2b}, we evaluate the impacts of $L_{dp}$ in \ref{sec:p2b}, $\alpha$ in \ref{sec:b2m} and $r$ in \ref{sec:FAF}.

\textbf{Impact of $L_{dp}$.} Fig.~\ref{fig: ablation}(a) showcases the influence of $L_{dp}$ on the quality of the pseudo masks. Compared to $L_{dp}$=0, the average IoU of the pseudo masks across the three datasets at $L_{dp}$=4 increased by approximately 2.5\%. The results indicate that extending perpendicular to the prediction direction aids in boundary estimation, as a more comprehensive target representation is necessary to effectively locate the target boundaries when facing irregularly shaped targets.

\textbf{Impact of $\alpha$.} Fig.~\ref{fig: ablation}(b) explores the effect of $\alpha$ on the quality of the pseudo masks and illustrates that selecting an appropriate weight $\alpha$ can yield an improvement of 3.5\% in the average IoU against that obtained with $\alpha$=0. This suggests that incorporating the pixel value of the point label into the pixel threshold enhances the accuracy of pseudo-mask generation.

\textbf{Impact of $r$.} Fig.~\ref{fig: ablation}(c) illustrates the impact of $r$ in FAF module on the filtered mask. Due to the maximum resolution of images in the datasets not exceeding 512, setting $r$=1000 results in the FAF module not filtering out any small targets. Consequently, the IoU of the filtered mask remains consistent with that of the prediction mask prior to filtering. In contrast, selecting an appropriate value for $r$ during the filtering process can lead to a significant improvement in IoU. When $r$=30, the average IoU across the three datasets increases by approximately 1.6\%, effectively filtering out most false detection small targets while preserving the detected small targets.

\begin{figure}[ht]
    \centering
    \begin{overpic}[width=\linewidth]{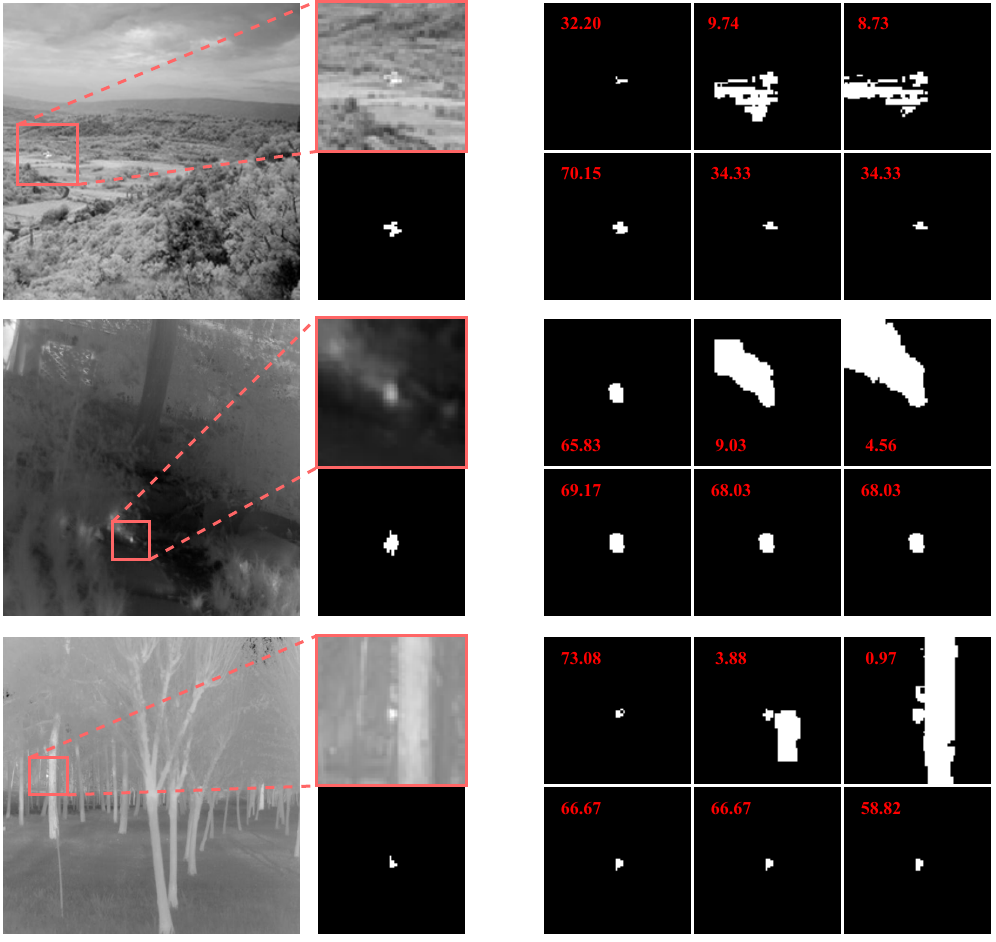} 
    \put(38,90){\footnotesize \textcolor{red}{IR}}
    \put(55,95){\footnotesize Size = 21}
    \put(70,95){\footnotesize Size = 41}
    \put(85,95){\footnotesize Size = 61}
    \put(38,75){\footnotesize \textcolor{green}{GT}}
    \put(38,58){\footnotesize \textcolor{red}{IR}}
    \put(38,43){\footnotesize \textcolor{green}{GT}}
    \put(38,26){\footnotesize \textcolor{red}{IR}}
    \put(38,11){\footnotesize \textcolor{green}{GT}}
    \put(51,80){\scriptsize \rotatebox{90}{MCLC \cite{li2023monte}}}
    \put(51,68){\footnotesize \rotatebox{90}{Ours}}
    \put(51,48){\scriptsize \rotatebox{90}{MCLC \cite{li2023monte}}}
    \put(51,36){\footnotesize \rotatebox{90}{Ours}}
    \put(51,16){\scriptsize \rotatebox{90}{MCLC \cite{li2023monte}}}
    \put(51,5){\footnotesize \rotatebox{90}{Ours}}
    \end{overpic}
    \caption{Qualitative comparison with different cropping sizes for MCLC and our PMG. The IoU between each pseudo-mask and the ground truth mask is in red.}
    \label{fig: crop_visual}
\end{figure}

\subsection{Visual Comparison}
\subsubsection{Comparison with Different Cropping sizes}
Fig.~\ref{fig: crop_visual} presents the visual results of the pseudo-masks generated by MCLC and PMG at different cropping sizes, along with the calculated IoU against the Ground Truth Masks. It is evident that as the cropping size increases, MCLC tends to produce false detection pixels, indicating its sensitivity to the chosen cropping size. In contrast, PMG consistently can generate reliable pseudo-masks, even when the cropping size significantly exceeds the actual target size.

\begin{figure*}[ht]
    \centering
    \begin{overpic}[width=\linewidth]{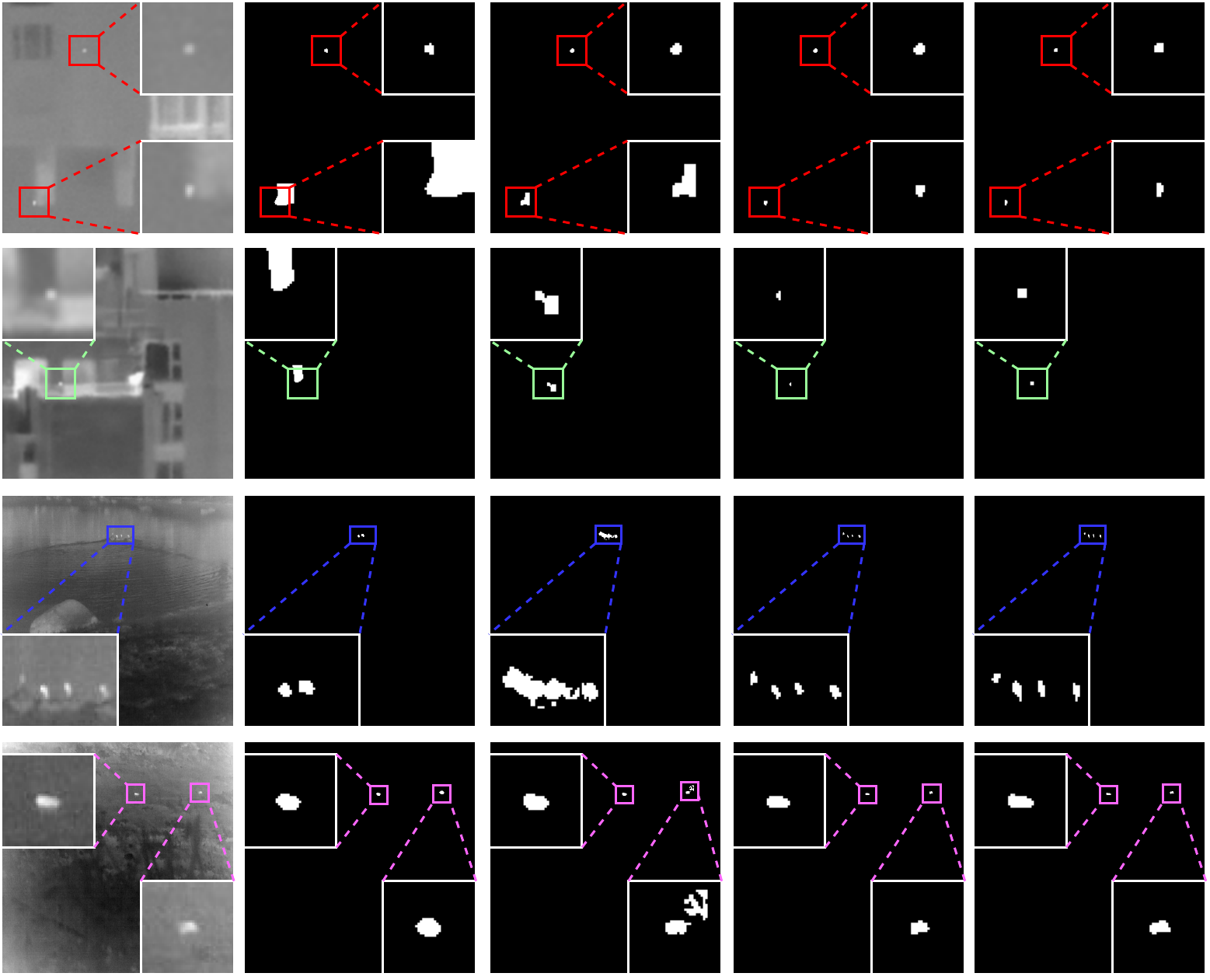}
    \put(5,-2){(a) IR Image}
    \put(24,-2){(b) COM \cite{li2024levelset}}
    \put(44,-2){(c) MCLC \cite{li2023monte}}
    \put(64,-2){(d) PMG(Ours)}
    \put(85,-2){(e) GT Mask}
    \end{overpic}
    \vspace{0.01cm}
    \caption{Visual examples of pseudo-masks generated by other learning-free pseudo-mask generation methods alongside our PMG.}
    \label{fig: visualization}
\end{figure*}

\subsubsection{Comparison with Different Methods}
Fig.~\ref{fig: visualization} provides a visual comparison of pseudo masks generated by COM \cite{li2024levelset} and MCLC \cite{li2023monte}, alongside initial masks generated by our PMG. The figure clearly shows that the other methods tend to produce false detection pixels when faced with background pixels that closely resemble the targets. In contrast, PMG generates pseudo masks that closely approximate the Ground Truth (GT) masks, demonstrating a higher level of accuracy.

\subsection{Further Analysis}
\subsubsection{Exploration in Real-World Scenrios}
We conducted extended experiments based on our method to address applications in real-world scenarios.

\textbf{With Coarse Centroid.} In practical situations, point annotations for small targets may not accurately correspond to the true centroid positions, but rather fluctuate around them. In our experiment, we refer to the coarse centroid annotation method from LESPS \cite{ying2023mapping}, using points that follow a Gaussian distribution around the centroid, with a standard deviation of 1/8 of the target size, as point labels. As shown in Table~\ref{tab:extended}, the decline in point label quality has minimal impact on our method, with even a slight increase in average IoU. We attribute this to two factors: first, our Points-to-Mask Generation method does not assume point labels as centroids, instead, it allows for multi-directional assessments of the pixels surrounding the point labels. Second, during the pseudo-mask updating process, the neural network's predictions operate independently of point labels, enabling the predictions to effectively correct the pseudo-masks.

\textbf{With Point Prompt.} In real-world scenarios, providing single-point annotations as a minimal-cost prompt to enhance detection performance also holds practical value. Therefore, we explored the detection performance when point labels are available during the inference process. Specifically, we followed the pseudo-mask updating process by first using PMG to obtain an initial mask based on the point label, and then integrating the initial mask with the model's predictions to generate a hybrid mask. As shown in Table~\ref{tab:extended}, when a single-point prompt is provided during testing, the average IoU increases by approximately 5.6\%, while maintaining $\rm P_d$ close to 100\%.

\begin{figure}[ht]
    \centering
    \begin{overpic}[width=\linewidth]{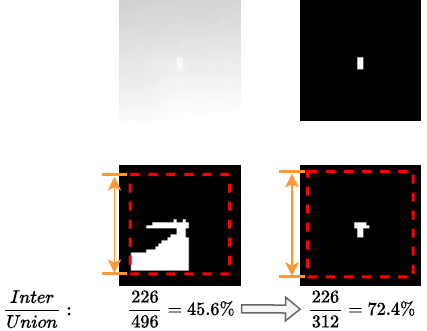}
    \put(32,43){(a) IR Image}
    \put(74,43){(b) GT Mask}
    \put(20,-5){(c) Initial masks with different cropping sizes}
    \put(21,24){33}
    \put(62,24){35}
    \end{overpic}
    \vspace{0.15cm}
    \caption{The example which leads to a sudden increase in IoU.}
    \label{fig: drop}
\end{figure}

\subsubsection{The Reason of Dramatic Rise in Fig.~\ref{fig: crop_target_type}(b)}
\label{sec:drop}
Upon examining the \textbf{Point} line in Fig.~\ref{fig: crop_target_type}(b), we observe that as the cropping size gradually increases, the IoU steadily declines. However, at a cropping size of approximately 33, there is a sudden increase in IoU. To investigate this unusual phenomenon, we conducted a detailed analysis.

Our findings indicate that in the example presented in Fig.~\ref{fig: drop}, the background in the lower left corner of the target exhibits minimal contrast with the target itself, resulting in a substantial number of false detection pixels in the pseudo mask generated at cropping size 33. At this cropping size, the IoU for the \textbf{Point} line across the three datasets is 226/496 = 45.6\%. In contrast, when the cropping size is increased to 35, a higher quality pseudo mask is produced for the target, effectively filtering out many of the false detection pixels; consequently, the IoU improves to 226/312 = 72.4\%, which accounts for the observed sudden increase in IoU.

This example illustrates that while a cropping size of 33 fully encompasses the small target, increasing the cropping size facilitates a more careful consideration of background pixels, which leads to the generation of higher-quality pseudo masks.

\begin{figure}[ht]
    \centering
    \begin{overpic}[width=\linewidth]{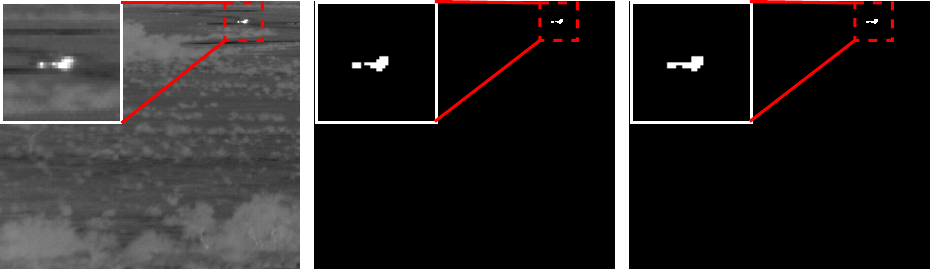}
    \put(5,-5){(a) IR Image}
    \put(38,-5){(b) GT Mask}
    \put(70,-5){(c) Initial Mask}
    \end{overpic}
    \vspace{0.01cm}
    \caption{An example of the "missed detection" phenomenon in the pseudo masks.}
    \label{fig: 100}
\end{figure}

\subsubsection{The Reason of Pd not Reaching 100\% in Table~\ref{tab:updating}}
\label{sec:100}
In Table~\ref{tab:updating}, both the initial mask and hybrid mask, which utilize point labels, are theoretically expected to encompass all small targets, yielding a Probability of detection ($\rm P_d$) of 100\%. However, the actual $\rm P_d$ falls short of this theoretical value.

Through a thorough investigation, we found that in the example illustrated in Fig.~\ref{fig: 100}, the Ground Truth (GT) mask shown in Fig.~\ref{fig: 100}(b) contains two adjacent small targets. However, the pseudo mask generated in Fig.~\ref{fig: 100}(c) may connect these two small targets, classifying them as a single target. This leads to a "missed detection" when calculating $\rm P_d$, ultimately preventing it from reaching the theoretical maximum of 100\%.

\section{Conclusion}
In this paper, we have proposed a simple but efficient Hybrid Mask Generation approach (HMG) to obtain high-quality pseudo masks from point labels by combining both strengths of deep-learning and learning-free strategies. Specifically, we design a learning-free Point-to-Mask Generation (PMG) method that shows robust adaptability without any prior information to recover initial masks from point labels and a Pseudo-Mask-Updating method that integrates deep-learning and learning-free strategies. Moreover, we found that our Point-to-Box method can flexibly integrate with other pseudo-mask generation approaches to reduce reliance on prior information, allowing for the effective fitting of infrared small targets of unknown sizes. Experimental results on three datasets demonstrated that our method overall outperforms other existing methods for infrared small target detection with single-point supervision.

% To print the credit authorship contribution details
% \printcredits

%% Loading bibliography style file
%\bibliographystyle{model1-num-names}
\bibliographystyle{cas-model2-names}

\begin{thebibliography}{10}

\bibitem{bearman2016whats}
Amy Bearman, Olga Russakovsky, Vittorio Ferrari, and Li~Fei-Fei.
\newblock What’s the point: Semantic segmentation with point supervision.
\newblock In {\em European Conference on Computer Vision}, pages 549--565, 2016.

\bibitem{bilen2016weakly}
Hakan Bilen and Andrea Vedaldi.
\newblock Weakly supervised deep detection networks.
\newblock In {\em Proceedings of the IEEE/CVF Conference on Computer Vision and Pattern Recognition}, pages 2846--2854, 2016.

\bibitem{chen2014local}
C.~L.~Philip Chen, Hong Li, Yantao Wei, Tian Xia, and Yuan~Yan Tang.
\newblock A local contrast method for small infrared target detection.
\newblock {\em IEEE Transactions on Geoscience and Remote Sensing}, 52(1):574--581, 2014.

\bibitem{chen2024tci-former}
Tianxiang Chen, Zhentao Tan, Qi~Chu, Yue Wu, Bin Liu, and Nenghai Yu.
\newblock Tci-former: Thermal conduction-inspired transformer for infrared small target detection.
\newblock In {\em Proceedings of the AAAI Conference on Artificial Intelligence}, 2024.

\bibitem{cheng2022pointly}
Bowen Cheng, Omkar Parkhi, and Alexander Kirillov.
\newblock Pointly-supervised instance segmentation.
\newblock In {\em Proceedings of the IEEE/CVF International Conference on Computer Vision}, pages 2607--2616, 2022.

\bibitem{ACM}
Yimian Dai, Yiquan Wu, Fei Zhou, and Kobus Barnard.
\newblock Asymmetric contextual modulation for infrared small target detection.
\newblock In {\em IEEE Winter Conference on Applications of Computer Vision}, pages 949--958, 2021.

\bibitem{ALCNet}
Yimian Dai, Yiquan Wu, Fei Zhou, and Kobus Barnard.
\newblock Attentional local contrast networks for infrared small target detection.
\newblock {\em IEEE Transactions on Geoscience and Remote Sensing}, 59(11):9813--9824, 2021.

\bibitem{d1999maxmean}
Suyog~D. Deshpande, Meng~Hwa Er, Ronda Venkateswarlu, and Philip Chan.
\newblock Max-mean and max-median filters for detection of small targets.
\newblock In {\em Optics \& Photonics}, 1999.

\bibitem{gao2013lowrank}
Chenqiang Gao, Deyu Meng, Yi~Yang, Yongtao Wang, Xiaofang Zhou, and Alexander~G. Hauptmann.
\newblock Infrared patch-image model for small target detection in a single image.
\newblock {\em IEEE Transactions on Image Processing}, 22(12):4996--5009, 2013.

\bibitem{Kingma2014AdamAM}
Diederik~P. Kingma and Jimmy Ba.
\newblock Adam: {A} method for stochastic optimization.
\newblock In {\em International Conference on Learning Representations}, 2015.

\bibitem{SAM}
Alexander Kirillov, Eric Mintun, Nikhila Ravi, Hanzi Mao, Chloe Rolland, Laura Gustafson, Tete Xiao, Spencer Whitehead, Alexander~C Berg, Wan-Yen Lo, et~al.
\newblock Segment anything.
\newblock In {\em ICCV}, pages 4015--4026, 2023.

\bibitem{kou2024mcgc}
Renke Kou, Chunping Wang, Qiang Fu, Zhanwu Li, Ying Luo, Boyang Li, Wei Li, and Zhenming Peng.
\newblock Mcgc: A multi-scale chain growth clustering algorithm for generating infrared small target mask under single-point supervision.
\newblock {\em IEEE Transactions on Geoscience and Remote Sensing}, 2024.

\bibitem{laradji2020proposal}
Issam~H Laradji, Negar Rostamzadeh, Pedro~O Pinheiro, David Vazquez, and Mark Schmidt.
\newblock Proposal-based instance segmentation with point supervision.
\newblock In {\em IEEE International Conference on Image Processing}, pages 2126--2130, 2020.

\bibitem{li2023monte}
Boyang Li, Yingqian Wang, Longguang Wang, Fei Zhang, Ting Liu, Zaiping Lin, Wei An, and Yulan Guo.
\newblock Monte carlo linear clustering with single-point supervision is enough for infrared small target detection.
\newblock In {\em Proceedings of the IEEE/CVF International Conference on Computer Vision}, pages 1009--1019, 2023.

\bibitem{DNANet}
Boyang Li, Chao Xiao, Longguang Wang, Yingqian Wang, Zaiping Lin, Miao Li, Wei An, and Yulan Guo.
\newblock Dense nested attention network for infrared small target detection.
\newblock {\em IEEE Transactions on Image Processing}, 32:1745--1758, 2022.

\bibitem{li2024first}
Boyang Li, Xinyi Ying, Ruojing Li, Yongxian Liu, Yangsi Shi, and Miao Li.
\newblock The first competition on resource-limited infrared small target detection challenge: Methods and results.
\newblock {\em arXiv preprint arXiv:2408.09615}, 2024.

\bibitem{li2024levelset}
Haoqing Li, Jinfu Yang, Yifei Xu, and Runshi Wang.
\newblock A level set annotation framework with single-point supervision for infrared small target detection.
\newblock {\em IEEE Signal Processing Letters}, 31:451--455, 2024.

\bibitem{li2021fully}
Yanwei Li, Hengshuang Zhao, Xiaojuan Qi, Liwei Wang, Zeming Li, Jian Sun, and Jiaya Jia.
\newblock Fully convolutional networks for panoptic segmentation.
\newblock In {\em Proceedings of the IEEE/CVF Conference on Computer Vision and Pattern Recognition}, pages 214--223, 2021.

\bibitem{liu2024cvpr}
Qiankun Liu, Rui Liu, Bolun Zheng, Hongkui Wang, and Ying Fu.
\newblock Infrared small target detection with scale and location sensitivity.
\newblock In {\em Proceedings of the IEEE/CVF Conference on Computer Vision and Pattern Recognition}, pages 17490--17499, 2024.

\bibitem{Paszke2019PyTorchAI}
Adam Paszke, Sam Gross, Francisco Massa, Adam Lerer, James Bradbury, Gregory Chanan, Trevor Killeen, Zeming Lin, Natalia Gimelshein, Luca Antiga, Alban Desmaison, Andreas K{\"{o}}pf, Edward~Z. Yang, Zachary DeVito, Martin Raison, Alykhan Tejani, Sasank Chilamkurthy, Benoit Steiner, Lu~Fang, Junjie Bai, and Soumith Chintala.
\newblock Pytorch: An imperative style, high-performance deep learning library.
\newblock In Hanna~M. Wallach, Hugo Larochelle, Alina Beygelzimer, Florence d'Alch{\'{e}}{-}Buc, Emily~B. Fox, and Roman Garnett, editors, {\em Annual Conference on Neural Information Processing Systems}, pages 8024--8035, 2019.

\bibitem{qian2019sceneparsing}
Rui Qian, Yunchao Wei, Honghui Shi, Jiachen Li, Jiaying Liu, and Thomas Huang.
\newblock Weakly supervised scene parsing with point-based distance metric learning.
\newblock In {\em Proceedings of the AAAI Conference on Artificial Intelligence}, pages 8843--8850, 2019.

\bibitem{Rawat2020ReviewOR}
Sur~Singh Rawat, Sashi~Kant Verma, and Yatindra Kumar.
\newblock Review on recent development in infrared small target detection algorithms.
\newblock {\em Procedia Computer Science}, 167:2496--2505, 2020.

\bibitem{Rivest1996DetectionOD}
Jean-François Rivest and Roger~A. Fortin.
\newblock Detection of dim targets in digital infrared imagery by morphological image processing.
\newblock {\em Optical Engineering}, 35:1886--1893, 1996.

\bibitem{MDvsFA}
Huan Wang, Luping Zhou, and Lei Wang.
\newblock Miss detection vs. false alarm: Adversarial learning for small object segmentation in infrared images.
\newblock In {\em Proceedings of the IEEE/CVF International Conference on Computer Vision}, pages 8508--8517, 2019.

\bibitem{wang2012infrared}
Xin Wang, Guofang Lv, and Lizhong Xu.
\newblock Infrared dim target detection based on visual attention.
\newblock {\em Infrared Physics \& Technology}, 55(6):513--521, 2012.

\bibitem{wei2017stc}
Yunchao Wei, Xiaodan Liang, Yunpeng Chen, Xiaohui Shen, Ming-Ming Cheng, Jiashi Feng, Yao Zhao, and Shuicheng Yan.
\newblock Stc: A simple to complex framework for weakly-supervised semantic segmentation.
\newblock {\em IEEE Transactions on Pattern Analysis and Machine Intelligence}, 39(11):2314--2320, 2017.

\bibitem{wu2022ship}
Peng Wu, Honghe Huang, Hanxiang Qian, Shaojing Su, Bei Sun, and Zhen Zuo.
\newblock Srcanet: Stacked residual coordinate attention network for infrared ship detection.
\newblock {\em IEEE Transactions on Geoscience and Remote Sensing}, 60:1--14, 2022.

\bibitem{yang2024EFLNet}
Bo~Yang, Xinyu Zhang, Jian Zhang, Jun Luo, Mingliang Zhou, and Yangjun Pi.
\newblock Eflnet: Enhancing feature learning network for infrared small target detection.
\newblock {\em IEEE Transactions on Geoscience and Remote Sensing}, 62:1--11, 2024.

\bibitem{yang2024label}
Dongning Yang, Haopeng Zhang, Ying Li, and Zhiguo Jiang.
\newblock Label evolution based on local contrast measure for single-point supervised infrared small target detection.
\newblock {\em IEEE Transactions on Geoscience and Remote Sensing}, 2024.

\bibitem{ying2023mapping}
Xinyi Ying, Li~Liu, Yingqian Wang, Ruojing Li, Nuo Chen, Zaiping Lin, Weidong Sheng, and Shilin Zhou.
\newblock Mapping degeneration meets label evolution: Learning infrared small target detection with single point supervision.
\newblock In {\em Proceedings of the IEEE/CVF Conference on Computer Vision and Pattern Recognition}, pages 15528--15538, 2023.

\bibitem{yuan2024beyond}
Shuai Yuan, Hanlin Qin, Renke Kou, Xiang Yan, Zechuan Li, Chenxu Peng, and Abd-Krim Seghouane.
\newblock Beyond full label: Single-point prompt for infrared small target label generation.
\newblock {\em arXiv preprint arXiv:2408.08191}, 2024.

\bibitem{yuan2024sctransnet}
Shuai Yuan, Hanlin Qin, Xiang Yan, Naveed Akhtar, and Ajmal Mian.
\newblock Sctransnet: Spatial-channel cross transformer network for infrared small target detection.
\newblock {\em IEEE Transactions on Geoscience and Remote Sensing}, 62:1–15, 2024.

\bibitem{empowering2021}
Jing Zhang and Dacheng Tao.
\newblock Empowering things with intelligence: A survey of the progress, challenges, and opportunities in artificial intelligence of things.
\newblock {\em IEEE Internet of Things Journal}, 8(10):7789--7817, 2021.

\bibitem{Zhang2024IRPruneDetEI}
Mingjin Zhang, Handi Yang, Jie Guo, Yunsong Li, Xinbo Gao, and Jing Zhang.
\newblock Irprunedet: Efficient infrared small target detection via wavelet structure-regularized soft channel pruning.
\newblock In {\em Proceedings of the AAAI Conference on Artificial Intelligence}, 2024.

\bibitem{Zhang2022ExploringFC}
Mingjin Zhang, Ke~Yue, Jing Zhang, Yunsong Li, and Xinbo Gao.
\newblock Exploring feature compensation and cross-level correlation for infrared small target detection.
\newblock {\em Proceedings of the 30th ACM International Conference on Multimedia}, 2022.

\bibitem{ISNet}
Mingjin Zhang, Rui Zhang, Yuxiang Yang, Haichen Bai, Jing Zhang, and Jie Guo.
\newblock Isnet: Shape matters for infrared small target detection.
\newblock In {\em Proceedings of the IEEE/CVF Conference on Computer Vision and Pattern Recognition}, pages 877--886, 2022.

\end{thebibliography}

% Loading bibliography database

% Biography
%\bio{}
% Here goes the biography details.
%\endbio

%\bio{pic1}
% Here goes the biography details.
%\endbio

\end{document}